\documentclass{ecai}

\usepackage{latexsym}
\usepackage{amssymb}
\usepackage{amsmath}
\usepackage{amsthm}
\usepackage{booktabs}
\usepackage{enumitem}
\usepackage{graphicx}
\usepackage{color}

\usepackage{adjustbox}
\usepackage[table,xcdraw]{xcolor}
\usepackage{makecell}
\usepackage{listings}
\usepackage{multirow}
\usepackage{tabularx,arydshln}
\usepackage{float}

\usepackage[justification=centering]{caption}

\usepackage{pgfplots}
\usepackage{pgfplotstable}
\usepgfplotslibrary{groupplots}
\usepgfplotslibrary{statistics}
\usetikzlibrary{matrix} 
\pgfplotsset{compat=newest}
\usetikzlibrary{pgfplots.fillbetween,patterns} 
\usepackage{tikz}
\usepackage{tcolorbox}

\definecolor{plotbackground}{HTML}{E6E6E6}

\newcommand\encircle[2][]{\tikz[overlay]\node[fill=blue!20,inner sep=2pt, anchor=text, rectangle, rounded corners=1.5mm,#1] {#2};\phantom{#2}}

\usepackage{xspace}
\usepackage{soul}      
\usepackage{microtype}
\newcommand{\feast}{\texttt{\textbf{FEAST}}\xspace}

\newcommand{\BibTeX}{B\kern-.05em{\sc i\kern-.025em b}\kern-.08em\TeX}

\usepackage{hyperref}

\usepackage{orcidlink}

\makeatletter
\def\blfootnote{\gdef\@thefnmark{*}\@footnotetext}
\makeatother

\begin{document}


\begin{frontmatter}


\paperid{2204} 


\title{\vspace{-0.5cm}\feast: Retrieval-Augmented Multi-Hierarchical Food Classification for the FoodEx2 System}


\author[A]{\fnms{Lorenzo}~\snm{Molfetta}\orcidlink{0009-0005-6817-6656}\footnote{Equal contribution (co-first authors).}}
\author[A]{\fnms{Alessio}~\snm{Cocchieri}\orcidlink{0009-0003-1507-1354}\footnotemark}
\author[A]{\fnms{Stefano}~\snm{Fantazzini}\orcidlink{0009-0006-2192-2604}\footnotemark}
\author[A]{\fnms{Giacomo}~\snm{Frisoni}\orcidlink{0000-0002-9845-0231}\footnotemark}
\author[A]{\fnms{Luca}~\snm{Ragazzi}\orcidlink{0000-0003-3574-9962}\footnotemark}
\author[A]{\fnms{Gianluca}~\snm{Moro}\orcidlink{0000-0002-3663-7877}\footnotemark}

\address[A]{Department of Computer Science and Engineering, University of Bologna, Cesena Campus,\\Via dell'Università 50, I-47522 Cesena, Italy}


\begin{abstract}
Hierarchical text classification (HTC) and extreme multi-label classification (XML) tasks face compounded challenges from complex label interdependencies, data sparsity, and extreme output dimensions.
These challenges are exemplified in the European Food Safety Authority's FoodEx2 system--a standardized food classification framework essential for food consumption monitoring and contaminant exposure assessment across Europe.
FoodEx2 coding transforms natural language food descriptions into a set of codes from multiple standardized hierarchies, but faces implementation barriers due to its complex structure.
Given a food description (e.g., ``organic yogurt''), the system identifies its base term (``yogurt''), all the applicable facet categories (e.g., ``production method''), and then, every relevant facet descriptors to each category (e.g., ``organic production'').
While existing models perform adequately on well-balanced and semantically dense hierarchies, no work has been applied on the practical constraints imposed by the FoodEx2 system.
The limited literature addressing such real-world scenarios further compounds these challenges.
We propose \feast (Food Embedding And Semantic Taxonomy), a novel retrieval-augmented framework that decomposes FoodEx2 classification into a three-stage approach: (1) base term identification, (2) multi-label facet prediction, and (3) facet descriptor assignment.
By leveraging the system's hierarchical structure to guide training and performing deep metric learning, \feast learns discriminative embeddings that mitigate data sparsity and improve generalization on rare and fine-grained labels.
Evaluated on the multilingual FoodEx2 benchmark, \feast outperforms the prior European's CNN baseline F1 scores by 12--38\% on rare classes. 
\end{abstract}

\end{frontmatter}

\blfootnote{The definitive, copyrighted, peer-reviewed, and edited version of this Article is published in \textit{ECAI 2025}, edited by I. Lynce et al., FAIA, pp. 4169–4176, 2025. DOI: \url{https://doi.org/10.3233/FAIA251309}.}


\section{Introduction}

Accurately mapping natural language (NL) descriptions to sparse hierarchical taxonomic structures in critical domains such as food safety systems overcomes the boundaries of a mere multi-label text classification task and transcends academic curiosity; it underpins effective public health measures, enforces regulatory rigor, and enables population-scale dietary analysis. 
In this context, Hierarchical Text Classification (HTC) and eXtreme Multi-Label Text Classification (XML) methods are employed to address specific, and often intertwined challenges.
While HTC is designed to exploit the set of labels $H$ organized according to a predefined hierarchical structure where the labels are connected in a parent-child relationship~\cite{DBLP:conf/acl/BanerjeeAPT19, DBLP:conf/eacl/JainRZYWGSZ24, DBLP:conf/emnlp/MaoTHR19}, XML is designed to manage the computational challenges inherent in datasets with a vast number of labels, aiming to identify a relevant subset of labels for each instance efficiently~\cite{DBLP:conf/nips/YouZWDMZ19, DBLP:conf/nips/ZhangCYD21}.
Despite their differences, both paradigms are integral to the real-world classification systems, where their combined strengths enable effective solutions.

\begin{figure}[t]
    \centering
    \includegraphics[width=\linewidth]{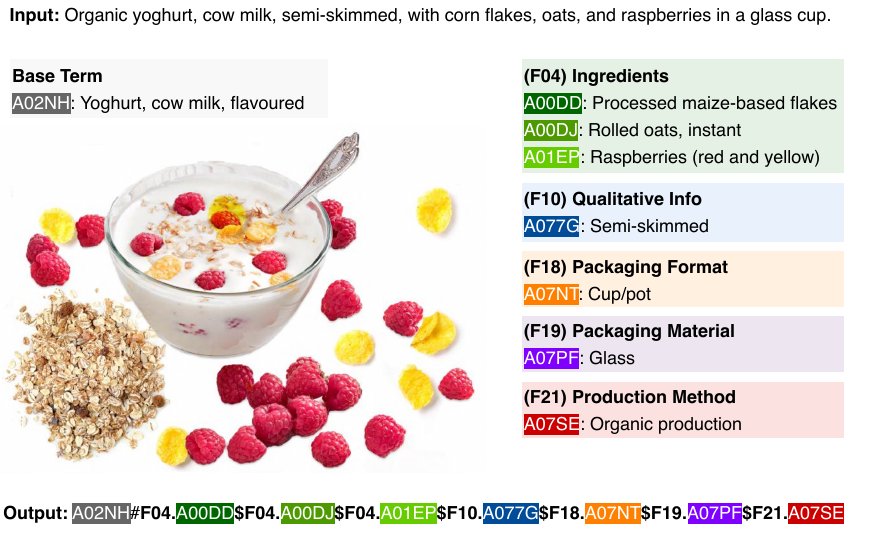}
    \caption{Illustration of the FoodEx2 coding system. An input food item is represented by a base term and facet descriptors detailing its key attributes, such as ingredients, processing, and packaging.}
    \vspace{10pt}
    \label{fig:input_output}
\end{figure}

With an increasing interest in efficient and sustainable AI~\cite{DBLP:conf/aaai/MoroRV23}, recent advances in HTC have primarily been driven by pre-trained transformers and graph-based techniques that integrate label semantics and hierarchical information into end-to-end models~\cite{DBLP:conf/acl/ChenMLY20, DBLP:conf/acl/WangWH0W22}.
Global architectures, which directly encode the entire label hierarchy in a single classifier, have demonstrated superior performance compared to local methods that deploy multiple individual classifiers for different parts of the hierarchy~\cite{DBLP:conf/acl/ZhouMLXDZXL20}. 
However, most of these models have been evaluated on well-balanced and semantically dense hierarchies.
Concerning XML, end-to-end trainable models that leverage pre-trained transformers have shown superior performance~\cite{DBLP:conf/aaai/JiangWSYZZ21}. 
Models such as XR-Transformer~\cite{DBLP:conf/nips/ZhangCYD21} and CascadeXML~\cite{DBLP:conf/nips/KharbandaBSB22} create hierarchical label trees through clustering algorithms to improve scalability, while AttentionXML~\cite{DBLP:conf/nips/YouZWDMZ19} employs multi-label attention mechanisms to incorporate contextual information and better predict rare labels.

In this paper, we examine FoodEx2, a food categorization system developed by the European Food Safety Authority (EFSA), which poses a far more realistic setting for HTC and XML research.
Originally designed to support tasks in food safety, nutrition, and contaminant exposure assessment, FoodEx2 encodes food items using a two-component structure: a unique base term and a series of facet categories with associated descriptors reporting key attributes of the food--as exemplified in Figure~\ref{fig:input_output}.
Unlike existing HTC and XML applications, FoodEx2 features interconnected ontological structures, where NL food descriptions must be mapped to codes belonging to multiple hierarchies containing thousands of elements.
Therefore, FoodEx2's food items simultaneously belong to various categorization dimensions (e.g., qualitative information, packaging information, production methods) across different hierarchical levels.
Furthermore, the sparse nature of the FoodEx2 label space raises several obstacles: (1) the uneven distribution of descriptors across facet categories requires models to handle extreme cases of class imbalance; (2) the facet codes must be derived solely from the NL input--despite the presence of extensive implicit facet information in the FoodEx2 taxonomy--accentuating the necessity for architectures that can effectively distill and exploit hierarchical, semantic, and contextual clues; (3) aligning the multi-faceted code generation with a structured taxonomy necessitates a system that not only classifies text accurately but also respects hierarchical dependencies during prediction.

To address these points, we introduce \textbf{\feast} (Food Embedding And Semantic Taxonomy), a novel framework that redefines the FoodEx2 classification problem as an HTC XML one.
By decomposing the problem into sub-tasks--base term identification, facet category determination, and facet descriptor classification--our solution effectively integrates hierarchical information across these elements while maintaining high classification accuracy.
Built on state-of-the-art encoder networks and large language models (LLMs) leveraging information from a hierarchical taxonomy, to the best of our knowledge, \feast represents the top-performing framework for food classification within the FoodEx2 system..


\section{Related Work}

\paragraph{Hierarchical Text Classification}
HTC methods generally fall into two main categories: local, which build separate classifiers at each level of the hierarchy, and global, which treat HTC as multi-label classification by integrating hierarchical structure directly into a single model~\cite{DBLP:conf/acl/BanerjeeAPT19,DBLP:conf/acl/ZhouMLXDZXL20}.
Modern HTC publications leverage transformer encoders such as BERT and RoBERTa to set strong baselines~\cite{DBLP:journals/corr/abs-2411-13687}, with distributed coordination mechanisms proving effective for managing complex hierarchical structures at scale~\cite{DBLP:conf/adc/LodiMS10,DBLP:journals/jnca/MoroM12}.
Exploiting semantic features through hybrid learning frameworks has shown significant improvements in text classification tasks~\cite{DBLP:conf/data/DomeniconiSLDKM16}.
Extensions like HiAGM~\cite{DBLP:conf/acl/ZhouMLXDZXL20} embed hierarchy via Tree‐LSTM or Graph Convolutional Networks (GCNs) to propagate label information, while HTCInfoMax~\cite{DBLP:conf/naacl/DengPHLY21} tackles label imbalance by enhancing representations of rare classes.
HiMatch~\cite{DBLP:conf/acl/ChenMLY20} further aligns document and label semantics using GCNs with hierarchy-aware matching losses.
Encoder-decoder models recast HTC as a generation task, with Seq2Tree~\cite{DBLP:conf/sigir/YuSM22} employing depth-first linearization of label trees to capture dependencies.

\paragraph{Extreme Multi‐Label Text Classification}

In XML, transformer-based methods have largely outperformed earlier bag-of-words and tree-based algorithms~\cite{DBLP:journals/corr/abs-2411-13687}.
XR-Transformer~\cite{DBLP:conf/nips/ZhangCYD21} introduces a recursive architecture with increasing label resolutions, forming a hierarchical label tree through k-means clustering on label features, while CascadeXML~\cite{DBLP:conf/nips/KharbandaBSB22} extends this by training a single transformer across multiple label resolutions, improving efficiency.
These architectures scale to hundreds of thousands of labels by pruning low-confidence branches and leveraging the label hierarchy to constrain predictions.
Traditional classifiers often lack flexibility when adding or modifying categories. 
However, procedures such as iterative refinement of category representations~\cite{DBLP:conf/ic3k/DomeniconiMPS14} and transfer learning for cross-domain adaptation~\cite{DBLP:conf/ic3k/DomeniconiMPP15} have shown promise in addressing these limitations.
Retrieval-augmented models address this issue by embedding queries and candidates into a shared semantic space~\cite{DBLP:conf/emnlp/LiZLGYTA23}.
Bi-encoder models generate dense embeddings for queries and candidates independently~\cite{DBLP:conf/emnlp/KarpukhinOMLWEC20}, while cross-encoders jointly encode query-label pairs for improved reranking at higher computational cost~\cite{DBLP:conf/emnlp/LiZLGYTA23}.

\paragraph{Applications to Food-Specific Domains}
Recent studies have addressed the classification of long texts in the food domain using hierarchical transformer-based models.
\cite{XIONG2023e17806} proposed a two-stage hierarchical transformer architecture to classify food safety news events. 
By chunking documents for BERT encoding, and then feeding embeddings into a transformer to capture inter-chunk context, this two-stage hierarchical design demonstrated improved accuracy in multi-class food safety event classification.
Although this work classified news event categories rather than a predefined food ontology like FoodEx2, it illustrates a hierarchical modeling technique tailored for handling long food-related descriptions.
\cite{article:Ozyegen-Classifying} investigates multi-level grocery products classification via dynamic masking to enforce hierarchical consistency during prediction. 
Their technique ensures that when the model predicts the parent category, they mask out any sub-categories that do not belong to one of the parents.
Their masking markedly improves hierarchical consistency and yields over 90\% F1 at the deepest level, compared to flat classifiers.

In food science, \cite{HU2023553} fine-tunes BERT to categorize packaged foods using ingredient lists and product names to assign category labels and predict nutrition quality scores.
They showed that transformers surpass rule-based methods and highlight the utility of contextual embeddings for mapping free-text to structured food categories.

Concerning hierarchy-aware formulations, \cite{narushynska2025enhancing} addresses product categorization within the GS1 Global Product Classification system, which includes a food hierarchy similar to FoodEx2.
By employing tree-based ensemble methods combined with a custom ``Penalized Information Gain'' metric that increases the weight of errors at higher hierarchy levels, they yielded a 12.7\% improvement in accuracy at the deepest level (and +12.6\% F1) compared to a flat classifier.
Although their method used gradient boosting trees rather than neural networks, their results underscore the benefit of incorporating and respecting hierarchical structures during model training.

While no papers target FoodEx2 coding system, its requirements align with other works: NL text processing~\cite{XIONG2023e17806}, shared contextual embedding spaces~\cite{HU2023553}, and strict parent-child dependencies~\cite{narushynska2025enhancing, article:Ozyegen-Classifying}.


\section{Data Curation}

The FoodEx2 dataset comprises 72,197 instances characterized by three features: the NL food description (\texttt{ENFOODNAME}), the FoodEx2 base term (\texttt{BASETERM\_NAME}), and the target FoodEx2 code (\texttt{FACETS}). 
The dataset employs a two-component structure: \textit{base terms} and \textit{facets}.
Since FoodEx2 does not provide the complete representation of base terms, facets, and auxiliary metadata, we used the latest available FoodEx2 Matrix catalog (MTX\_12).

\paragraph{Base Term}
Primary food item identifier.
Each description must be associated with precisely one base term (4,367 possible ones).

\paragraph{Facets}
Facets are categorized into explicit and implicit ones.
Explicit facets are modifiers directly mentioned in the NL input, while implicit facets are intrinsic properties of the base term.
There are a total of 28 facet categories, each one possessing a specific number of facet descriptors, ranging from as few as 2 (for facet \texttt{F26}) to as many as 19,744 (for facet \texttt{F01}). 

\paragraph{Code Format}
The encoding follows a specific syntax where the base term is followed (optionally) by ``\#'' and a sequence of facet categories with related facet descriptors, each group separated by ``\$''.
Figure~\ref{fig:input_output} shows an example used by the FoodEx2 system.

\begin{figure*}[t]
    \centering   
    \includegraphics[width=\linewidth]{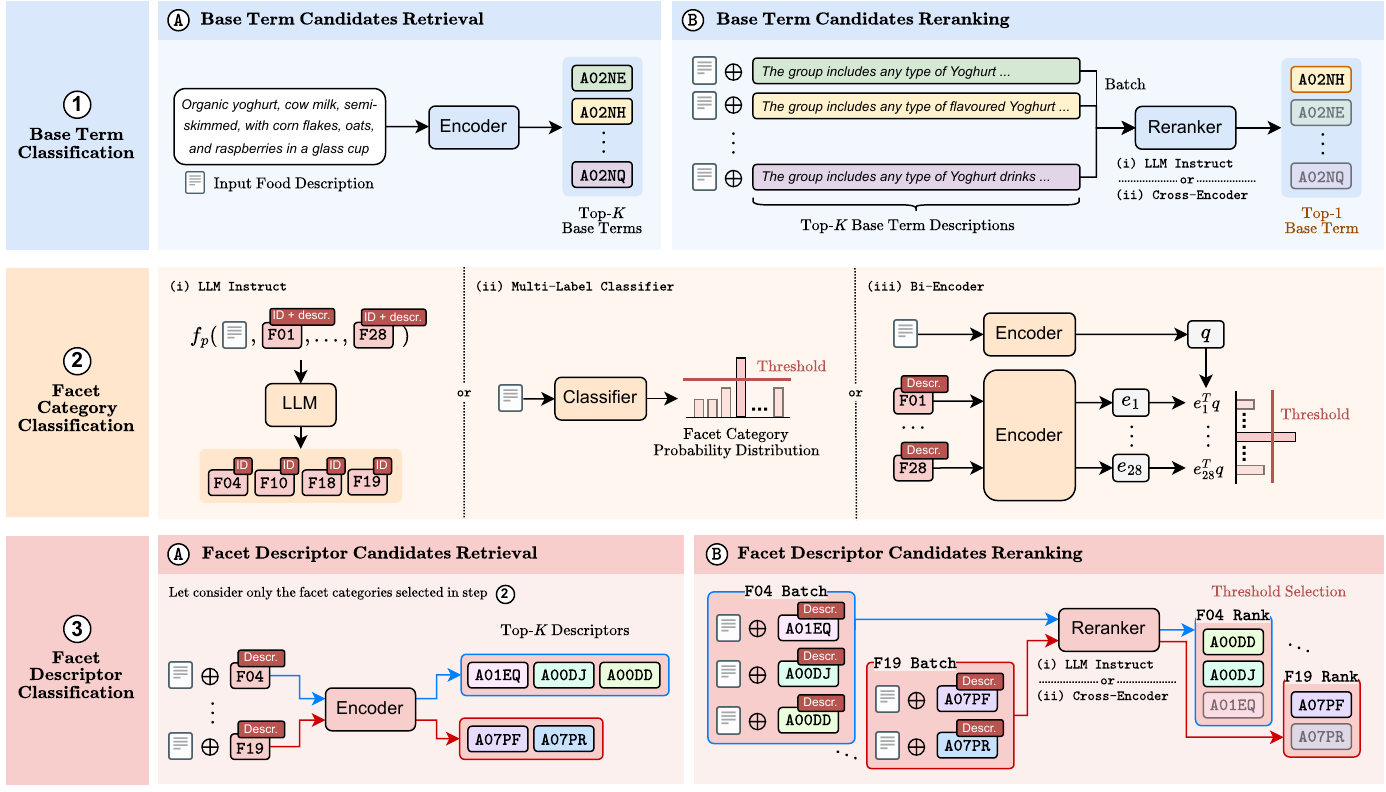}
    \caption{Overview of the proposed FoodEx2 food classification pipeline, which operates in three main stages: (1) \textbf{Base Term Classification}, where top-$K$ candidate base terms are retrieved and reranked to produce a top-1 base term; (2) \textbf{Facet Category Classification}, which identifies relevant facet categories using either an LLM, a multi-label classifier, or a bi-encoder approach; and (3) \textbf{Facet Descriptor Classification}, which retrieves and reranks descriptor candidates for the previously selected facet categories. The system combines encoder-based retrieval with flexible reranking strategies (LLM instruct or cross-encoder) to map free-text food descriptions into structured FoodEx2 codes. $\oplus$ denotes concatenation and $f_p(\cdot)$ represents a prompt template function.}
    \label{fig:method}
    
\end{figure*}

\subsection{Preprocessing}

We performed extensive preprocessing to ensure data consistency and reliability.
To this end, we removed entries with missing values (198), duplicates (30,314), and food descriptions with anonymized information (4,536).
For the remaining entries, we standardized inconsistent input formats, corrected malformed facet codes, resolved character encoding issues, and addressed cases where identical descriptions were mapped to multiple target codes by retaining only the most informative annotation.
Starting from 72,197 entries, these steps reduced the dataset to 28,648.

\subsection{Key Observations}
The dataset's class coverage before and after preprocessing can be seen in Table~\ref{tab:class_coverage}, while distributions for facet's categories and descriptions are reported in Table~\ref{tab:facet_distribution}.
The FoodEx2 taxonomy presents several intricate characteristics that impact classification performance.

\begin{table}[t]
\caption{Coverage of base terms and facets in our dataset relative to those available in MTX\_12, both before and after preprocessing.}
\centering
\small
\begin{adjustbox}{max width=\linewidth}
\begin{tabular}{lccccc}
\toprule
& \multicolumn{2}{c}{\textbf{Before Preprocessing}} & \multicolumn{2}{c}{\textbf{After Preprocessing}} \\
\cmidrule(lr){2-3} \cmidrule(lr){4-5}
\textbf{Metric} & \textbf{Coverage} & \textbf{\%} & \textbf{Coverage} & \textbf{\%} \\
\midrule
Base term class & 1,726 / 4,367 & 39.52 & 1,650 / 4,367 & 37.78 \\
Facet category & 27 / 28 & 96.43 & 28 / 28 & 100.00 \\
Facet descriptor & 1,508 / 28,675 & 5.26 & 1,355 / 28,675 & 4.73 \\
\bottomrule
\end{tabular}
\end{adjustbox}
\label{tab:class_coverage}
\end{table}

\paragraph{Multilingualism}
The dataset exhibits multilingual content, with many entries incorporating specialized culinary terms or food references from various languages (i.e., ``\textit{Rabbit, leg, cooked, preservation method undefined, intérieur cuit/à point}'').

\paragraph{Taxonomy Semantics}
We observed non-intuitive hierarchical relationships where semantically similar items often appear in distant positions within the taxonomy. 
For instance, ``\textit{Custard}'' is positioned as the only base term at the same level as ``\textit{Starchy pudding},'' yet it is taxonomically distant from ``\textit{Rice pudding}'' despite their conceptual similarities. 
This non-intuitive arrangement suggests successful classification models must capture deeper semantic relationships between food items rather than relying on surface-level pattern recognition.

\paragraph{Expert Knowledge Requirements}
Many target classes require mapping that is not obvious from surface-level names, demanding what we term ``abstractive linkage''. 
For example, the input ``\textit{caramel peanut butter nougat roasted peanuts covered with milk chocolate}'' is expected to be associated with the facet ``\textit{White sugar}'' \texttt{[A032J]}, a connection that requires domain-specific knowledge about food composition and ingredients.  
This highlights the need for solutions that can incorporate implicit knowledge and make inferential connections beyond the explicit content of food descriptions.

\begin{table}[t]
\caption{Distribution of facet categories and descriptors per instance after data preprocessing.}
\centering
\small
\begin{adjustbox}{max width=\linewidth}
\begin{tabular}{lccccccc}
\toprule
& \multicolumn{3}{c}{\textbf{Facet Categories}} & \multicolumn{3}{c}{\textbf{Facet Descriptors}} & \\
\cmidrule(lr){2-4} \cmidrule(lr){5-7}
\textbf{Split} & \textbf{Min} & \textbf{Avg} & \textbf{Max} & \textbf{Min} & \textbf{Avg} & \textbf{Max} & \textbf{Base Term-only} \\
\midrule
Train & 0 & 1.75 & 9 & 0 & 2.23 & 17 & 2,796 \\
Eval  & 0 & 1.59 & 6 & 0 & 2.08 & 16 & 213 \\
Test  & 0 & 1.69 & 7 & 0 & 2.15 & 11 & 58 \\
\bottomrule
\end{tabular}
\end{adjustbox}
\label{tab:facet_distribution}
\end{table}

\subsection{Data Splitting}

To establish evaluation splits, we employed stratified sampling to maintain consistent class distributions across sets.
Our test set contains 991 samples, while our validation set consists of 926 samples.
To accelerate validation and minimize costs, we simulated a set of distractor candidate classes as if they were retrieved. 
Distractors are selected following the same technique described in Section~\ref{subsec:negMin}. Their use was confined to the validation set, while the complete real pipeline, involving all components of our methodology, was limited to the test set.
To assess generalization to unseen categories and reduce the risk of overfitting, we constructed an out-of-sample (OOS) evaluation set. 
This set was built by grouping samples based on their base term identifier and selecting a subset of groups to form the test portion, ensuring disjoint categories between training and test. 
Groups were added incrementally until the test set reached approximately 1,000 entries. 
The remaining groups were used as the training set.
Unless otherwise specified, all experimental results reported in this paper were obtained using the OOS dataset.	


\section{Methodology}
\label{sec:methodology}

We outline the methodological foundations on which \texttt{FEAST} is built.
\subsection{Hard Negative Mining}
\label{subsec:negMin}

To improve the model’s ability to resolve fine-grained semantic distinctions, we developed hybrid data selection combining taxonomy-based and semantic similarity-based strategies, ensuring the model distinguishes fine-grained differences between closely related terms.

\paragraph{Taxonomy-Based Sampling}

Leveraging its hierarchical nature, we can represent the FoodEx2 ecosystem as a tree structure $\mathcal{T} = (V, E)$, where vertices correspond to unique terms and edges denote parent-child relationships. 
For each target term $t \in V$, we identified hard negatives through a two-phase process:\\
(1) \textit{Candidate Selection:} we selected the set of nodes $C_{pa}$ sharing the same parent $\textit{pa}(v)$ as the target term $t$:
\begin{equation}
\label{eq:cpa}
C_{pa}(t) = \{ v \in V : \textit{pa}(v) = \textit{pa}(t), v \neq t \}
\end{equation}
(2) \textit{Hierarchical Structured Selection:} when insufficient candidates were available, we sampled additional nodes based on structural proximity and semantic overlap. Proximity was computed using a hop distance $d_{\text{hop}}$ based on the Lowest Common Ancestor (LCA):
\begin{equation}
\label{eq:dhop}    
\begin{split}
z &= \text{LCA}(t, v) \\
d_{\text{hop}}(t, v) &= \lvert \textit{dp}(t) - \textit{dp}(z) \rvert + \lvert \textit{dp}(v) - \textit{dp}(z) \rvert
\end{split}
\end{equation}
with $\textit{dp}(\cdot)$ being the node depth in the tree, with $\textit{dp}(\texttt{ROOT})=0$.

Candidates are further ranked by semantic overlap based on the \texttt{ImplicitFacets} of the target node $t$ and candidate node $v$.
Letting $\mathcal{F}(v)$ be the set of implicit facets of the candidate node $v$, we define the semantic overlap as the number of shared implicit facets:
\begin{equation}
o_{\text{facets}}(t,v) = |\mathcal{F}(t) \cap \mathcal{F}(v)|
\end{equation}
The resulting scoring function used to sample candidates is:
\begin{equation}
\mathcal{S}(t,v) = \frac{1 + o_{\text{facets}}(t,v)}{1 + d_{\text{hop}}(t,v)}
\end{equation}
Candidates are then sampled probabilistically according to their normalized scores as follows:
\begin{equation}
\label{eq:tax-sampl}
p(v|t) = \frac{\mathcal{S}(t,v)}{\sum_{u \in \mathcal{T}} \mathcal{S}(t,u)}
\end{equation}
Thus, given $\mathbf{N} = \lvert \mathcal{H}_{\text{tax}}(t) \rvert$, the set of taxonomy-based hard negatives, formally denoted as $\mathcal{H}_{\text{tax}}(t)$, is obtained as:
\begin{equation}
\mathcal{H}_{\text{tax}}(t) = C_{pa}(t) \cup \{ v_i \sim p(v|t)\}_{\mathbf{N} - \lvert C_{pa}(t) \rvert} 
\end{equation}

\subsection{Retrieval-Enhanced Classification}
\label{subsec:retrieval_class}

Traditional classifiers assign labels to inputs based on lexical features, often failing to capture semantic nuances such as synonymy, polysemy, or domain-specific terminology.
To this end, we propose a retrieval-enhanced framework designed to map textual inputs into dense semantic representations. 

\paragraph{Model Architecture}

For the retrieval process, we employ a bi-encoder architecture with shared parameters between the query and document encoding branches, following state-of-the-art practices~\cite{DBLP:conf/emnlp/ReimersG19}.
A transformer encoder $\mathcal{E}$ processes an input sequence $x$ into contextualized token representations $\mathbf{H}$ as follows:
\begin{equation}
    \mathbf{H} = \mathcal{E}_{\theta}(x), \quad \mathbf{H} \in \mathbb{R}^{n \times h}
\end{equation}
where $n$ denotes the number of tokens in the input sequence, and $h$ the hidden dimension of the encoder.
Successively, a mean pooling operation generates a unified embedding.

\paragraph{Training Objective}

The training objective utilizes the Multiple Negative Ranking Loss (MNR)~\citep{DBLP:journals/corr/HendersonASSLGK17}, which minimizes the semantic distance between query-positive document pairs while maximizing the distance to negative examples.
For a batch $\mathcal{B}$ of query-document pairs $(q_i, d_i)$, with the set of explicitly mined hard negatives $\mathcal{H}(q_i)$, the MNR Loss function is defined as follows:
\begin{equation}
\label{eq:mnrLoss}
\begin{split}
\mathcal{L}_\text{RETR}&(q, d; \theta, \psi) = -\frac{1}{\mathcal{B}}\sum_{i=1}^{\mathcal{B}}  \Biggl[ \,\text{sim}(q_i, d_i) \, + \\ 
&- \log\left(e^{\,\text{sim}(q_i, d_i)} + \sum_{d_j \in \mathcal{H}(q_i)} e^{\,\text{sim}(q_i, d_j)}\right)\Biggr]
\end{split}
\end{equation}
where cosine similarity is computed between the embeddings.

\subsection{Entailment-Based Document Reranking}
\label{subsec:reranking}

Although embedding-based retrieval reduces irrelevant candidates, subtle semantic differences can still lead to false positives among similar retrieved terms. We address this through a dedicated reranking module that reassesses the top-ranked candidates.

Given an input query $q$ and the set of retrieved top-$k$ document description $\mathcal{D}_k^{(q)} = \{ d_1, d_2, \dots, d_k \}$, the reranking model $\mathcal{R}_{\phi}$ assigns each input-candidate pair a relevance score. 
Unlike the bi-encoder used for initial retrieval, we employ a cross-encoder that processes concatenated query-candidate pairs $(q, d_i)$ through a transformer encoder $\mathcal{E}$ followed by a pooling layer $\mathcal{P}$ and linear projection:
\begin{equation}
     \rho_\phi(q, d_i) = \mathbf{w}^\top \cdot \mathcal{P}\circ\mathcal{E}([q; d_i]) + b
\end{equation}
where $\mathbf{w}$ and $b$ are trainable parameters, and ``;'' is the concatenation operation.
Training leverages hard negatives mined via the strategy in Section~\ref{subsec:negMin}, using binary cross-entropy loss to distinguish positive pairs (aligned query-candidate pairs) from negatives.
During inference, each candidate is independently assessed by the trained cross-encoder.
We apply a confidence threshold $\tau$ to filter candidates further, retaining only those with sufficiently high relevance scores.

\begin{table*}[ht!]
    \caption{Retrieval and reranking scores (\%). \textbf{Bold} and \underline{underlined} values indicate the best and second-best result for each metric. \hspace{0.5pt} \encircle[fill=blue!15, text=black]{Blue} \hspace{0.5pt} highlights the best performing retrieval model, while \hspace{0.5pt} \encircle[fill=orange!15, text=black]{orange} \hspace{0.5pt} highlights the best performing reranker in each task averaging over all metrics.}
    \centering
    \begin{adjustbox}{max width=\linewidth}
    \begin{tabular}{>{\arraybackslash}p{1.5cm}llccccccccc} 
        \toprule
        \textbf{Stage} & \textbf{Task} & \textbf{Model} & \textbf{Acc@1} & \textbf{Acc@3} & \textbf{Acc@5} & \textbf{Rec@1} & \textbf{Rec@3} & \textbf{Rec@5} & \textbf{NDCG@10} & \textbf{MRR@10} & \textbf{MAP@100} \\
        \midrule
        \multirow{8}{=}{\textbf{Retriever}} & \multirow{4}{*}{\textbf{Base Term (I)}} 
        & GTE-Multilingual-base & 95.10 & 99.71 & 99.89 & 94.84 & 99.60 & 99.87 & 98.01 & 97.33 & 97.30\\
        && BGE-M3 & \textbf{95.95} & \underline{99.82} & \textbf{99.93} & \underline{95.67} & \underline{99.70} & \underline{99.91} & \underline{98.38} & \underline{97.83} & \underline{97.81}\\
        && EuroBERT-610M & 95.10 & 99.78 & 99.90 & 95.20 & 99.68 & 99.89 & 98.29 & 97.80 & 97.77 \\
        && \cellcolor{blue!15}{ModernBERT} & \cellcolor{blue!15}{\underline{96.57}} & \cellcolor{blue!15}{\textbf{99.82}} & \cellcolor{blue!15}{\underline{99.92}} & \cellcolor{blue!15}{\textbf{96.31}} & \cellcolor{blue!15}{\textbf{99.72}} & \cellcolor{blue!15}{\textbf{99.91}} & \cellcolor{blue!15}{\textbf{98.63}} & \cellcolor{blue!15}{\textbf{98.16}} & \cellcolor{blue!15}{\textbf{98.14}} \\
        \cline{2-12}
        & \multirow{4}{*}{\textbf{Descriptor (III)}} 
        & GTE-Multilingual-base & 98.37 & 99.90 & 99.98 & 86.88 & 99.21 & 99.84 & 99.27 & 99.13 & \textbf{99.94} \\
        && BGE-M3 & \underline{98.76} & \underline{99.92} & \textbf{99.98} & \underline{87.21} & 99.28 & \underline{99.85} & \underline{99.45} & \underline{99.33} & 98.21\\
        && EuroBERT-610M & 98.60 & 99.91 & 99.97 & 87.05 & \underline{99.30} & 99.84 & \underline{99.45} & 99.31 & 99.28 \\
        && \cellcolor{blue!15}{ModernBERT} & \cellcolor{blue!15}{\textbf{98.90}} & \cellcolor{blue!15}{\textbf{99.93}} & \cellcolor{blue!15}{99.97} & \cellcolor{blue!15}{\textbf{87.32}} & \cellcolor{blue!15}{\textbf{99.40}} & \cellcolor{blue!15}{\textbf{99.87}} & \cellcolor{blue!15}{\textbf{99.53}} & \cellcolor{blue!15}{\textbf{99.41}} & \cellcolor{blue!15}{\underline{99.33}} \\
        \hdashline[2pt/2pt]
        \multirow{6}{=}{\textbf{Reranking}} & \multirow{3}{*}{\textbf{Base Term (I)}} 
        & DeBERTa-v3-small & \underline{81.01} & \underline{89.69} & \underline{90.00} & \underline{89.81} & \underline{90.00} & \underline{99.72} & \underline{85.12} & \underline{86.38} & \underline{85.12}\\
        && DeBERTa-v3-base & 76.01 & 84.69 & 85.00 & 87.06 & 89.93 & 98.80 & 90.12 & 81.38 & 80.12 \\
        && \cellcolor{orange!15}{DeBERTa-v3-large} & \cellcolor{orange!15}{\textbf{91.01}} & \cellcolor{orange!15}{\textbf{99.69}} & \cellcolor{orange!15}{\textbf{100.0}} & \cellcolor{orange!15}{\textbf{94.31}} & \cellcolor{orange!15}{\textbf{100.0}} & \cellcolor{orange!15}{\textbf{99.96}} & \cellcolor{orange!15}{\textbf{95.12}} & \cellcolor{orange!15}{\textbf{96.38}} & \cellcolor{orange!15}{\textbf{95.12}} \\
        \cline{2-12}
        & \multirow{3}{*}{\textbf{Descriptor (III)}} 
        & DeBERTa-v3-small & \underline{86.03} & \underline{89.57} & \underline{90.00} & \underline{85.13} & \underline{89.74} & \underline{90.82} & \underline{87.80} & \underline{88.17} & \underline{87.35}\\
         && DeBERTa-v3-base & 81.03 & 84.57 & 85.00 & 82.70 & 84.74 & 94.82 & 82.80 & 83.17 & 82.35\\
        && \cellcolor{orange!15}{DeBERTa-v3-large} & \cellcolor{orange!15}{\textbf{96.03}} & \cellcolor{orange!15}{\textbf{99.57}} & \cellcolor{orange!15}{\textbf{100.0}} & \cellcolor{orange!15}{\textbf{96.70}} & \cellcolor{orange!15}{\textbf{99.74}} & \cellcolor{orange!15}{\textbf{99.82}} & \cellcolor{orange!15}{\textbf{97.80}} & \cellcolor{orange!15}{\textbf{98.17}} & \cellcolor{orange!15}{\textbf{97.35}} \\
        \bottomrule
    \end{tabular}
    \end{adjustbox}
    \vspace{4pt}
    \label{tab:combined_results}
\end{table*}

\subsection{LLM for Classification}

For the classification phase, we adopt a flexible approach that allows choosing between a lightweight encoder-only reranking model and a more expressive generative LLM. While encoder-only models are computationally efficient and fast, generative LLMs offer greater versatility through multi-instruction tuning, enabling them to handle diverse classification tasks under a unified model. 
This capability facilitates knowledge sharing across tasks, potentially improving generalization, though at a higher computational cost.

We fine-tune an LLM jointly on the three classification subtasks, framing classification as a sequence generation problem. 
Figure~\ref{fig:prompts} illustrates the prompt templates used for instruction tuning, guiding the model to predict base terms, categories, or descriptors by leveraging NL context and candidate options.

\begin{figure}[t!]
    \centering
    \small
    \begin{tcolorbox}[colframe=black, colback=gray!5, title=Prompt Templates for LLM Instruction Tuning]
    \textbf{Task I: Base Term Classification} \\
    \texttt{system:} ``Given a food item, select one or more base terms from the candidate options. The context provides additional information for each candidate.'' \\
    \texttt{user:} ``Context:\{context\} BaseTerms: \{baseterms\} Food: \{food\}''
    
    \medskip
    
    \textbf{Task II: Facet Category Classification} \\
    \texttt{system:} ``Given a food item, select one or more categories that best describe it. The context provides additional information for each candidate. If no category applies, return an empty list [].'' \\
    \texttt{user:} ``Context: \{context\} Categories: \{categories\} Food: \{food\}''
    
    \medskip
    
    \textbf{Task III: Facet Descriptor Identification} \\
    \texttt{system:} ``Given a food item in a category, select its most relevant descriptor(s) from the candidate options. The context provides additional information for each candidate descriptor.'' \\
    \texttt{user:} ``Context: \{context\}  Descriptors: \{descriptors\} Food: \{food\} Category: \{food\_category\}''
    \end{tcolorbox}
    \caption{Prompt templates used for LLM's multitask instruction tuning across base term, category, and descriptor classification tasks.}
    \label{fig:prompts}
    \vspace{10pt}
\end{figure}

\section{\feast}
\label{sec:implementation}
The proposed \feast framework approaches the FoodEx2 classification task through a multi-phase prediction process.
Specifically, we decompose the task into three sequential sub-tasks: base term classification, facet category classification, and facet descriptor identification.
Figure~\ref{fig:method} provides a graphical summary.

\subsection{Task I - Base Term Classification}

Given an input NL food description \texttt{inputText}, we identify its most relevant FoodEx2 base term through three modules.

\paragraph{Base Term Retriever}
We leverage base term descriptions as context to disambiguate relevant base terms from user input. 
We compute embedding representations of the food description and retrieve the top-$k$ most similar base terms:
\begin{equation}
    \label{eq:baseTermRetr}
    \mathcal{D}_k^{(q)} = \mathcal{M}(\texttt{inputText})
\end{equation}

\paragraph{Base Term Reranker}
To refine retrieval results and address subtle semantic differences, we implement a reranking model that assigns relevance scores to each candidate pair and filters results using a confidence threshold $\tau$:
\begin{equation}
    \overline{\mathcal{D}}_{\text{rank}}(\texttt{inputText}) = \mathcal{R}(\texttt{inputText}, \mathcal{D}_k^{(q)})
\end{equation}

\paragraph{LLM-Based Base Term Classifier}
Alternatively, we employ a fine-tuned LLM as a single-step text classifier, framing classification as a text-generation task guided by NL instructions.

\subsection{Task II - Facet Category Classification}

To identify applicable facet categories, we analyze three alternatives.

\paragraph{Multi-Label Classifier}
An encoder is fine-tuned for multi-label text classification. Given input text, the model produces probability scores for each facet category $i \in \{1, \dots, N\}$, applying a threshold $\tau$ to determine binary assignments: $\hat{y}_i = 1$ if $\sigma(z_i) \ge \tau$ else $0$. The model is trained by minimizing the Binary Cross-Entropy loss.

\paragraph{Bi-Encoder Classifier}
Two independent encoders generate embeddings for the concatenation $t_i = \texttt{inputText}_i \bigoplus \texttt{baseterm}_i$ and for each facet category description, respectively. 
We compute similarity scores between these embeddings, refine them through a multilayer perceptron, and apply a threshold to determine relevant categories.

\paragraph{LLM-Based Classifier}
Similar to the base term classifier, an LLM is prompted with the input description and the list of candidate facets. The model then generates the most relevant facet category.

\subsection{Task III - Facet Descriptor Identification}
We identify appropriate descriptors for each identified facet category.

\paragraph{Descriptor Retriever}
For each input text and associated facet category, we retrieve the top-$k$ most relevant descriptors:
\begin{equation}
    \label{eq:retrPh3}
\mathcal{D}_k = \mathcal{M}(\texttt{inputText}, \text{descr}(c))
\end{equation}

\paragraph{Descriptor Reranker}
We refine the retrieved descriptors using a reranker that evaluates contextual relevance:
\begin{equation}
    \label{eq:rankph3}
\overline{\mathcal{D}}_{\text{ranked}} = \mathcal{R}(\texttt{inputText}, \text{descr}(c), \mathcal{D}_k)
\end{equation}

\paragraph{LLM-Based Descriptor Classifier}
As with previous tasks, we also implement a single-step generative approach where an LLM generates relevant descriptors for each category from a set of candidates.

\section{Experimental Setup}

\subsection{Metrics}
Our system evaluation employs two complementary metric categories.
For retrieval performance, we utilize Accuracy@K, Recall@K, Precision@K, Mean Reciprocal Rank (MRR), and Mean Average Precision (MAP) to assess both the ranking quality and the system's ability to surface relevant items effectively.
For classification, we employ Precision, Recall, Support, and both Micro and Macro-averaged F1 scores to capture the system's labeling accuracy and coverage across potentially imbalanced class distributions.
These metrics are used to evaluate each sub-task component of our pipeline.

\subsection{Hyperparameters}

Below, we highlight the architecture and training configurations.

\paragraph{Task I} 
For retrieval, we used the Sentence Transformer models with backbones including BGE-M3~\cite{BAAI_bge-m3}, GTE-multilingual-base~\cite{zhang2024mgte}, ModernBERT-Large~\cite{DBLP:journals/corr/abs-2412-13663}, and EuroBERT-610M~\cite{boizard2025eurobertscalingmultilingualencoders}, trained with a maximum sequence length of 64 and cosine similarity loss. The rerankers employed DeBERTa-v3~\cite{DBLP:conf/iclr/HeLGC21} (small/base/large) with a sequence length of 256 and binary cross-entropy loss.

\paragraph{Task II} 
For classification, we used DeBERTa-v3~\cite{DBLP:conf/iclr/HeLGC21} (small/base) and RoBERTa~\cite{DBLP:journals/corr/abs-1907-11692} trained for 3 epochs, with 256 max sequence length, batch size 8, $2e{-}5$ learning rate, and linear decay. 

\paragraph{Task III} 
We again employed a retrieve-and-rerank approach following the same approach as in Task 1, but with sequence length increased to 96 and batch sizes appropriate to the different input characteristics.

\paragraph{Models}
We set the reproducibility random seed to 42 and used the AdamW optimizer with hyper-parameters $\beta_1$=$0.9$, $\beta_2$=$0.999$.
For the LLM-based classifier, we fine-tuned LLaMA-3.1-8B~\cite{meta2024llama31} jointly on all three tasks using LoRA adapters with rank $r$=$32$ and $\alpha$=$32$. We set a max sequence length of 3072 and trained for 1 epoch with a batch size of 32, a learning rate of $2e{-}4$, and a cosine decay scheduler.

\subsection{Hardware Configuration}
Each run, except for the LLM fine-tuning, was performed on an internal workstation using a single Nvidia GeForce RTX3090 GPU with 24~GB of dedicated memory, 64~GB of RAM, and an Intel® Core™ i9-10900X (3.70GHz) CPU. For LLMs, we used an NVIDIA A100 with 80~GB of dedicated memory, 128~GB of RAM, and an Intel® Core™ i9-10900X (3.70GHz) CPU. The reference operating system is Ubuntu 20.04.3 LTS.


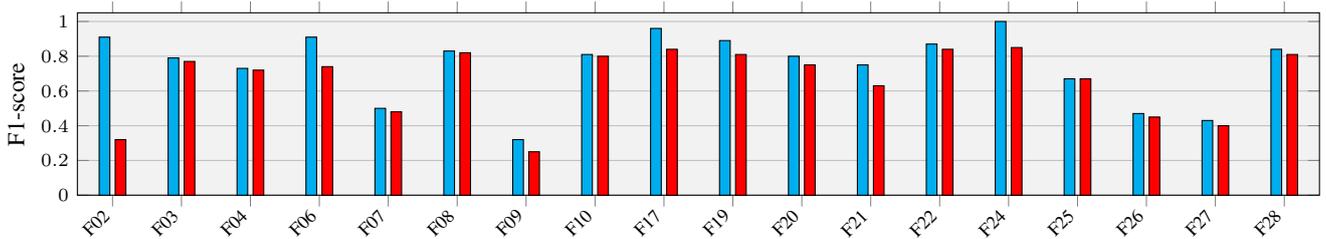
\begin{figure*}[t]
  \centering
  \begin{tikzpicture}
   \begin{axis}[
    ybar,
    symbolic x coords={F02,F03,F04,F06,F07,F08,F09,F10,F17,F19,F20,F21,F22,F24,F25,F26,F27,F28},
    xtick=data,
    x tick label style={rotate=45, anchor=east},
    ymin=0, ymax=1.05,
    ylabel={F1-score},
    legend pos=north east,
    width=\textwidth,
    ymajorgrids=true,
    ytick={0,0.2,0.4,0.6,0.8,1.0},
    yminorgrids=false,
    height=4cm,
    bar width=4pt, 
    enlarge x limits=0.03,
    axis background/.style={fill=plotbackground!50},
    every tick label/.append style={font=\fontsize{7}{7}\selectfont},
   ]
    \addplot[fill=cyan] coordinates {(F02,0.91) (F03,0.79) (F04,0.73) (F06,0.91) (F07,0.50) (F08,0.83) (F09,0.32) (F10,0.81) (F17,0.96)  (F19,0.89) (F20,0.80) (F21,0.75) (F22,0.87) (F24,1.00) (F25,0.67) (F26,0.47) (F27,0.43) (F28,0.84)};
    \addplot[fill=red] coordinates {(F02,0.32) (F03,0.77) (F04,0.72) (F06,0.74) (F07,0.48) (F08,0.82) (F09,0.25) (F10,0.80) (F17,0.84)  (F19,0.81) (F20,0.75) (F21,0.63) (F22,0.84) (F24,0.85) (F25,0.67) (F26,0.45) (F27,0.40) (F28,0.81)};
   \end{axis}
  \end{tikzpicture}
  \caption{Comparison of F1-scores between \hspace{0.5pt} \encircle[fill=cyan, text=white]{LLaMA-3.1-8B} \hspace{0.5pt} \citep{meta2024llama31} and \hspace{0.5pt} \encircle[fill=red, text=white]{DeBERTa-v3-small} \hspace{0.5pt} \citep{DBLP:conf/iclr/HeLGC21} across different facet categories (Task II).}
  \label{fig:f1_comparison}
 \end{figure*}

\section{Results}
\label{sec:results}

We evaluate our pipeline components across multiple tasks. Performance metrics for each module are presented below.

\paragraph{Retrievers}

Our fine-tuned ModernBERT retriever proves strong performance on both base term and descriptor retrieval tasks (see Table~\ref{tab:combined_results}).
In the Base Term task, it delivers 96.57\% Acc@1, NDCG@10 of 98.63\%, MRR@10 of 98.16\%, and MAP@100 of 98.14\%. 
For Descriptor retrieval, it achieves 98.90\% Acc@1, NDCG@10 of 99.53\%, MRR@10 of 99.41\%, and MAP@100 of 99.33\%. 
Across all intermediate thresholds (Acc@3/5 and Rec@1/3/5), ModernBERT consistently outperforms the competing models.

\paragraph{Rerankers}
As shown in Table~\ref{tab:combined_results}, the reranker refines retrieval quality. For base terms, it achieves 91.01\% accuracy@1, with NDCG@10 at 95.12\% and MRR@10 at 96.38\%. For facet descriptors, the reranker obtains an accuracy@1 of 96.03\%, and similarly robust ranking metrics (NDCG@10 = 97.80\%, MRR@10 = 98.17\%). These results show the reranker's capacity to enhance ranking precision.

\paragraph{Bi-Encoder Classifier}

Results are given in Table~\ref{tab:biencoder_res} and Table~\ref{tab:bi-encoder_micro_macro}. The bi-encoder classifier shows moderate effectiveness with 73.03\% accuracy@1. While accuracy improves at higher ranks (accuracy@3 = 76.46\%, accuracy@5 = 77.37\%, accuracy@10 = 78.28\%), ranking quality remains a challenge. 
Precision@1 reaches 73.03\% but declines at broader ranks (precision@10 = 16.58\%), with recall@10 at 77.46\%.

\begin{table}[t]
    \caption{Bi-encoder classifier results on facet category classification (Task II); b and s denote base and small size.}
    \centering
    \begin{adjustbox}{max width=\linewidth}
    \begin{tabular}{lcccccccc}
        \toprule
        & \multicolumn{4}{c}{\textbf{Accuracy} (\%)} & \multicolumn{4}{c}{\textbf{Recall} (\%)}
        \\
        \cmidrule(lr){2-5}
        \cmidrule(lr){6-9}
        \textbf{Model} & \textbf{@1} & \textbf{@3} & \textbf{@5} & \textbf{@10} & \textbf{@1} & \textbf{@3} & \textbf{@5} & \textbf{@10} \\
        \midrule
        RoBERTa-b & 51.72 & 62.93 & 65.76 & 71.21 & 29.28 & 42.61 & 50.73 & 62.30 \\
        DeBERTA-v3-s & 71.11 & 76.46 & \textbf{77.58} & 77.88 & 43.47 & 68.94 & 74.26 & 76.34 \\
        \rowcolor{lightgray!20} \textbf{DeBERTA-v3-b} & \textbf{73.03} & \textbf{76.46} & 77.37 & \textbf{78.28} & \textbf{44.70} & \textbf{70.01} & \textbf{75.14} & \textbf{77.46} \\
        \bottomrule
    \end{tabular}
    \end{adjustbox}
    \vspace{4pt}
    \label{tab:biencoder_res}
\end{table}

\begin{table}[t]
    \caption{Bi-encoder classifier with DeBERTa-v3-base performance metrics across different thresholds for both Macro and Micro averaging for Task II.}
    \centering
    \small
    \begin{adjustbox}{max width=\linewidth}
    \begin{tabular}{lccccccc}
        \toprule
         & \multicolumn{4}{c}{\textbf{Macro}} & \multicolumn{3}{c}{\textbf{Micro}} \\
        \cmidrule(lr){2-5} \cmidrule(lr){6-8}
         \textbf{Threshold} & \textbf{Acc} & \textbf{Prec} & \textbf{Rec} & \textbf{F1} & \textbf{Prec} & \textbf{Rec} & \textbf{F1} \\
        \midrule
        0.4 & 72.93 & 57.57 & 50.62 & 53.05 & 89.52 & 86.52 & 88.00 \\
        0.35 & 73.13 & 57.00 & 52.58 & 54.00 & 88.25 & 87.30 & 87.77 \\
        0.3 & 71.72 & 55.53 & 53.55 & 53.81 & 86.06 & 88.02 & 87.03 \\
        \bottomrule
    \end{tabular}
    \end{adjustbox}
    \vspace{4pt}
    \label{tab:bi-encoder_micro_macro}
\end{table}

\begin{table}[!t]
   \caption{LLaMA-3.1-8B performance results across tasks on the corresponding validation sets. Exact Match is computed based on exact matches of the full set of labels in multi-label classification. Precision, recall, and F1-score are reported as micro-averages.}
    \centering
    \small
    \begin{adjustbox}{max width=\linewidth}
    \begin{tabular}{lccccc}
        \toprule
        \textbf{Task} & \textbf{Acc} (\%) & \textbf{EM} (\%) & \textbf{Prec} (\%) & \textbf{Rec} (\%) & \textbf{F1} (\%)\\
        \midrule
        Task I & 99.99 & 98.75 & 99.59 & 98.78 & 99.18 \\
        Task II & 96.39 & 55.65 & 83.25 & 74.83 & 78.82 \\
        Task III & 99.99 & 99.54 & 100.00 & 99.64 & 99.82 \\
        \bottomrule
    \end{tabular}
    \end{adjustbox}
    \vspace{4pt}
    \label{tab:llm_results}
\end{table}

\paragraph{Multi-Label Classifier}
Despite severe class imbalance and limited training data, our supervised classifier demonstrates robust performance while using only 142M parameters (see Figure~\ref{fig:f1_comparison}).
Even relatively rare classes (i.e., F06, F20, F21, F24) achieve high F1 scores, with only the most underrepresented classes showing lower F1 scores.

\paragraph{Large Language Model}
Table~\ref{tab:llm_results} reveals a performance disparity in our fine-tuned LLM, which excels at Tasks I and III but delivers suboptimal results for Task II.
The model’s performance on several facet categories for Task II underperforms when compared to a simpler classifier model.
This degradation occurs because facet category classification requires identifying multiple, potentially overlapping semantic dimensions simultaneously from a single food description, a challenge better addressed through the continuous semantic space of dense embeddings rather than discrete pattern matching in LLM prompts. 
In contrast, Tasks I and III involve more direct semantic matching that aligns well with the LLM's contextual pattern recognition capabilities within structured prompts.


\section{Discussion}

Our experiments show that \feast effectively tackles the FoodEx2 coding challenge by using retrieval-augmented, hierarchical modeling.
The results confirm that combining deep metric learning with the FoodEx2 taxonomy helps overcome data sparsity and improves generalization to fine-grained labels.
The current European standard for food classification relies on a collection of CNN networks~\cite{livaniou2024efsa}. 
These networks achieve average accuracy metrics of 83\%, 87\%, and 87\% on the three tasks, respectively.
Since our top-performing retriever model achieves a perfect 100\% recall score in the @10 setting, we can use the reranker's Acc@1 as our benchmark when providing the 10 retrieved elements as input.
Under this assumption, \feast achieves an accuracy of 91.01\%, 96.39\%, and 96.03\% on the three tasks, respectively,  thus standing as the state-of-the-art model in the food classification domain in the FoodEx2 domain.

Nevertheless, the limited coverage of the available FoodEx2 tender dataset constrains the number of available base terms and the diversity of facets.
Class imbalance remains a challenge, with underrepresented facet categories exhibiting lower classification performance, even achieving a zero F1 score. 
Augmenting the training corpus through targeted sampling, synthetic data generation (e.g., prompt-based descriptor synthesis), or hierarchical regularization could further enhance model robustness on low-support classes.
Although our test metrics suggest strong potential, they derive from the reduced tender dataset, which may not fully reflect the real-world classification scenario of FoodEx2. 
The lack of standardized public benchmarks in this domain also limits comparative analysis with alternative approaches.


\section{Conclusion and  Future Work}

We proposed \feast, a retrieval-augmented, hierarchical framework for automated FoodEx2 coding that integrates dense retrieval, cross-encoder reranking, and multi-label classification. 
Beyond strong empirical results, \feast's modular design offers both interpretability--since retrieved candidates and reranking scores at each stage can be inspected by experts--and scalability, as the retrieval-based design allows seamless updates when new codes or facets emerge, minimizing the need for full-model retraining.
Our evaluations demonstrate that \feast significantly outperforms the previous European CNN-based baseline, particularly excelling in handling rare and fine-grained classes where existing models often struggle. 
By leveraging deep metric learning and hierarchical modeling, \feast effectively addresses data sparsity and class imbalance--two major challenges inherent to the FoodEx2 taxonomy.
Despite these advancements, the absence of a publicly standardized benchmark for FoodEx2 coding limits broad comparative analysis.
Looking forward, we plan to develop an end-to-end pipeline that jointly optimizes base term identification, facet prediction, and descriptor assignment, improving consistency and harnessing inter-task dependencies.
Further, incorporating data augmentation strategies and continual learning mechanisms could enhance \feast's robustness and adaptability to evolving taxonomies.


\section*{Ethical Statement}

The raw dataset and model weights used in this research are currently not publicly available due to pending approval from the Tender's contracting authority.
We anticipate making these resources accessible to the research community once we receive the necessary clearance from the relevant authorities.

\begin{ack}
Research partially supported by: \href{https://aipact-edih.it}{AI-PACT} (CUP B47H22004450008, B47H22004460001); National Plan PNC-I.1 \href{https://www.fondazionedare.it/en/progetto-obiettivi-struttura/}{DARE} (PNC0000002, CUP B53C22006450001); PNRR Extended Partnership \href{https://fondazione-fair.it/en/}{FAIR} (PE00000013, Spoke 8); 2024 Scientific Research and High Technology Program, project ``\href{https://disi-unibo-nlp.github.io/projects/carisbo/}{AI analysis for risk assessment of empty lymph nodes in endometrial cancer surgery}'', the Fondazione Cassa di Risparmio in Bologna; Chips JU \href{https://tristan-project.eu/team/}{TRISTAN} (G.A. 101095947). LG Solution Srl for partially funding a PhD scholarship to L. Molfetta. Dinova Srl for partially funding a PhD scholarship to S. Fantazzini.
\end{ack}


\bibliography{bibliography}


\end{document}